\documentclass[conference]{IEEEtran}
\IEEEoverridecommandlockouts
\usepackage{cite}
\usepackage{amsmath,amssymb,amsfonts}
\usepackage{bm}
\usepackage{textcomp}
\usepackage{xcolor}
\usepackage{tikz}
\usetikzlibrary{shapes, arrows, positioning, calc, fit}
\usepackage{amsmath}
\usepackage[T1]{fontenc}
\usepackage{graphicx}
\usepackage{cite}
\usepackage{algpseudocode}
\usepackage{amsmath,amssymb,amsfonts}
\usepackage{amsthm}
\usepackage{textcomp}
\usepackage{xcolor}
\usepackage{subcaption}
\usepackage{upgreek}
\usepackage{booktabs}
\newcommand{\mycomment}[1]{}
\usepackage{balance}
\usepackage[hidelinks]{hyperref}
\def\BibTeX{{\rm B\kern-.05em{\sc i\kern-.025em b}\kern-.08em
    T\kern-.1667em\lower.7ex\hbox{E}\kern-.125emX}}

\makeatletter
\newcommand{\linebreakand}{%
  \end{@IEEEauthorhalign}
  \hfill\mbox{}\par
  \mbox{}\hfill\begin{@IEEEauthorhalign}
}

\newcommand{\ddtheta}{\frac{d}{d\theta}}

\newcommand{\Normal}{\mathcal{N}}
\newcommand{\logN}{\log\mathcal{N}}

\newcommand{\nx}{{n_X}}

\newcommand{\resample}{\kappa}

\usepackage[ruled,vlined,linesnumbered]{algorithm2e}
\newcommand{\resampleu}{\nu} 

\begin{document}

\title{Enhanced SMC$^2$: Leveraging Gradient Information from Differentiable Particle Filters Within Langevin Proposals\\
\thanks{This research was funded in whole, or in part, by the Wellcome Trust [226691/Z/22/Z]. For the purpose of Open Access, the author has applied a CC BY public copyright license to any Author Accepted Manuscript version arising from this submission. CR was funded by the Wellcome CAMO-Net UK grant: 226691/Z/22/Z; JM was funded by a Research Studentship jointly funded by the EPSRC Centre for Doctoral Training in Distributed Algorithms EP/S023445/1; AV, PH and SM were funded by EPSRC through the Big Hypotheses under Grant EP/R018537/1}
\thanks{\bf{\footnotesize \textcopyright 2024 IEEE. Personal use of this material is permitted. Permission from IEEE must be obtained for all other uses, in any current or future media, including reprinting/republishing this material for advertising or promotional purposes, creating new collective works, for resale or redistribution to servers or lists, or reuse of any copyrighted component of this work in other works.}}
}

\author{\IEEEauthorblockN{Conor Rosato\IEEEauthorrefmark{1}, Joshua Murphy\IEEEauthorrefmark{2}, Alessandro Varsi\IEEEauthorrefmark{2}, Paul Horridge\IEEEauthorrefmark{2}
and
Simon Maskell\IEEEauthorrefmark{2}}\\
\IEEEauthorblockA{\IEEEauthorrefmark{1} Department of Pharmacology and Therapeutics, University of Liverpool, United Kingdom \\
\IEEEauthorrefmark{2} Department of Electrical Engineering and Electronics, University of Liverpool, United Kingdom\\
Email: \{cmrosa, joshua.murphy, a.varsi, p.horridge, smaskell\}@liverpool.ac.uk}}


\maketitle

\begin{abstract}

Sequential Monte Carlo Squared (SMC$^2$) is a Bayesian method which can infer the states and parameters of non-linear, non-Gaussian state-space models. The standard random-walk proposal in SMC$^2$ faces challenges, particularly with high-dimensional parameter spaces. This study outlines a novel approach by harnessing first-order gradients derived from a Common Random Numbers - Particle Filter (CRN-PF) using PyTorch. The resulting gradients can be leveraged within a Langevin proposal without accept/reject. Including Langevin dynamics within the proposal can result in a higher effective sample size and more accurate parameter estimates when compared with the random-walk. The resulting algorithm is parallelized on distributed memory using Message Passing Interface (MPI) and runs in $\mathcal{O}(\log_2N)$ time complexity. Utilizing 64 computational cores we obtain a 51x speed-up when compared to a single core. A GitHub link is given which provides access to the code.

\end{abstract}


\section{Introduction}\label{sec:introduction}

Markov Chain Monte Carlo (MCMC) algorithms that use gradient information relating to the target distribution have been well studied and are widely used in practice. Examples of such proposals include the Metropolis-Adjusted Langevin Algorithm (MALA) \cite{roberts1996exponential}, Hamiltonian Monte Carlo (HMC) \cite{neal2012mcmc} and the No-U-Turn Sampler (NUTS) \cite{NUTS}. Probabilistic programming languages such as Stan \cite{carpenter2017stan} and PyMC4 \cite{pymc3} allow users to define statistical models and perform inference automatically using NUTS. Recent advancements in the context of Sequential Monte Carlo Samplers (SMC) \cite{del2006sequential} with HMC and NUTS proposals can be found in \cite{smchmc} and \cite{devlin2021no}, respectively. Sequential Monte Carlo Squared (SMC$^2$) \cite{chopin2013smc2} and Particle- Markov Chain Monte Carlo (p-MCMC) \cite{andrieu_doucet_holenstein_2010} are two Bayesian methods that use a Particle Filter (PF)  \cite{arulampalam2002tutorial} to estimate the parameters and dynamic states of State-Space Models (SSMs). SMC$^2$ is a relatively mature algorithm but widely regarded as prohibitively time consuming in applied contexts. Recent work in \cite{rosato2023mathcal} outlines a SMC$^2$ framework which is implemented across distributed memory architectures which has an optimized L-kernel that runs in $\mathcal{O}(\log_2N)$ time-complexity. Results indicate that with 128 processors, a $54\times$ speed-up can be seen when compared to p-MCMC. Utilizing gradient information within SMC$^2$ is yet to be considered. This is due to the gradient of the log-likelihood from a PF needing to be calculated. It has been noted in literature that the PF contains inherent operations that are non-differentiable \cite{rosato2022efficient, corenflos2021differentiable}. Therefore, the current state-of-the-art in terms SMC$^2$ is the Random-Walk (RW) proposal which suffers from the \textit{curse of dimensionality}.

A recent review outlines numerous methods to differentiate the PF \cite{chen2023overview} with advantages and disadvantages for each of the methods defined. Applications of differentiable PFs include: parameter estimation in SSMs \cite{rosato2022efficient, rosato2023disease}, robot localization \cite{corenflos2021differentiable, karkus2018particle}, neural networks \cite{cox2024end, li2024revisiting} and autonomous driving \cite{singh2023particle}. Reference \cite{poyiadjis2011particle} provides two methods to estimate the gradient of the log-likelihood and the observed information matrices at $\bm{\theta}$ using PF methods. As the number of particles within the PF increases, the first method has a computational cost that grows quadratically while the second computes estimates whose variance increases quadratically. A method for differentiating this variant of the PF in PyTorch \cite{paszke2019pytorch} is provided in \cite{diffresam_1}. The work presented in \cite{poyiadjis2011particle} has formed the basis of parameter estimation of SSMs using p-MCMC with Langevin dynamics \cite{particleLangevin, inproceedings} and p-MCMC with first- and second-order Hessian information \cite{Second-order, Dahlin_2014}. 

An alternative method for estimating the gradients from a Common Random Numbers - Particle Filter (CRN-PF) that takes into account all samples from all iterations is presented in \cite{rosato2022efficient}. Reference \cite{rosato2022efficient} also shows how the resulting gradients can be used within p-MCMC with HMC and NUTS proposals. Empirical results are provided that show the approach outlined in \cite{rosato2022efficient} is competitive in terms of run-time and estimation accuracy when compared to other differentiable PFs. 

A limitation of differentiating the PF with CRN, that has been highlighted in the publication \cite{arya2022automatic}, is that the gradient evaluations in \eqref{eq:first_gradientlogposterior} will be biased for any one random number seed. We overcome this issue by assigning each particle in the SMC sampler, at each iteration, a different seed. Therefore, estimates are calculated as averages over multiple seeds. This addresses concerns related the bias inherent in the approach described in \cite{rosato2022efficient} where estimates were calculated using one common seed for all particles at all iterations.

The contribution of this paper is as follows: we utilize and extend the parallel SMC$^2$ framework in \cite{rosato2023mathcal} to include first-order gradients estimated from a CRN-PF, using PyTorch's automatic differentiation, within a Langevin proposal. 
  
The structure of this paper is as follows: the CRN-PF is outlined in Section~\ref{sec:particle_filter}. The methodology for calculating the log-likelihood and first-order gradients are presented in Sections~\ref{sec:pf_likelihood} and \ref{sec:pf_gradient} respectively. The distributed SMC$^2$ framework is described in Section~\ref{sec:SMC_sampler} with the various proposals represented in Section~\ref{sec:proposals}. This is followed by numerical results and conclusions and future work in Sections~\ref{sec:Examples} and \ref{sec:conclusions}, respectively.



\section{Particle Filter}\label{sec:particle_filter}

State-Space Models (SSMs) have been used to represent the dependence of latent states in non-linear non-Gaussian dynamical systems in a broad-spectrum of research fields \cite{doucet2001sequential}. A SSM consists of a state equation, 
\begin{equation}\label{xt}
\mathbf{x}_{t} \mid \mathbf{x}_{t-1} \sim f(\mathbf{x}_{t} \mid \mathbf{x}_{t-1}, \bm{\theta}),
\end{equation}
\noindent and an observation equation 
\begin{equation}\label{yt}
\mathbf{y}_{t} \mid \mathbf{x}_{t} \sim g(\mathbf{y}_{t} \mid \mathbf{x}_{t}, \bm{\theta}),
\end{equation}
\noindent which are parameterized by $\bm{\theta}$ which has $D$ dimensions: $\bm{\theta}=\{\theta_1, \theta_2, \dots, \theta_D\}$. 

Consider a SSM that for $t$ timesteps simulates the states, $x_{0:t}=\{x_0, x_1,\dots, x_t\}$, and obtains data at each increment of time $y_{0:t}=\{y_0, y_1,\dots, y_t\}$. In filtering problems, an estimate of the posterior distribution over the latent states can be made recursively via prediction and update steps using the observable data $p(\mathbf{x}_{0:t}|\mathbf{y}_{1:t})$. If the Markov process is of order one, the posterior can be defined as $p(\mathbf{x}_{t}|\mathbf{y}_{1:t})$ such that the posterior from the previous time step and \eqref{xt} can be used to predict the latent state $x_t$ using the Chapman-Kolmogorov equation
\begin{align}
p(\mathbf{x}_{t}|\mathbf{y}_{1:t-1}) =& \int p(\mathbf{x}_t|\mathbf{x}_{t-1},\mathbf{y}_{1:t-1})p(\mathbf{x}_{t-1}|\mathbf{y}_{1:t-1})dx_{t-1}\nonumber\\ 
=& \int p(\mathbf{x}_t|\mathbf{x}_{t-1})p(\mathbf{x}_{t-1}|\mathbf{y}_{1:t-1})dx_{t-1}\label{eq:Chapman-Kolmogrov},
\end{align}
where $p(\mathbf{x}_t|\mathbf{x}_{t-1})$ is equal to $p(\mathbf{x}_t|\mathbf{x}_{t-1},\mathbf{y}_{1:t-1})$. Bayes' theorem can be used to perform the update step by defining the posterior as 
\begin{align}\label{eq:filtering_bayes_}
p(\mathbf{x}_{t}|\mathbf{y}_{1:t}) &=\frac{p(\mathbf{y}_{t}|\mathbf{x}_{t})p(\mathbf{x}_{t}|\mathbf{y}_{1:t-1})}{p(\mathbf{y}_t|\mathbf{y}_{1:t-1})} 
\nonumber\\&=\frac{p(\mathbf{y}_t|\mathbf{x}_t)p(\mathbf{x}_{t}|\mathbf{y}_{1:t-1})}{\int p(\mathbf{y}_t|\mathbf{x}_t)p(\mathbf{x}_{t}|\mathbf{y}_{1:t-1})dx_{t}}.
\end{align}
In high dimensions, the integrals in \eqref{eq:Chapman-Kolmogrov} and \eqref{eq:filtering_bayes_} can become impossible to solve. The PF can recursively approximate these integrals using a set of $N_x$ particles (samples) and principles adopted from importance sampling. At $t=0$, a set of $N_x$ particles are drawn from a proposal distribution $q\left(\mathbf{x}_{1}\right)$ and given a weight of $1/N_x$ such that the set of $N_x$ particles is defined as $\left\{\mathbf{x}_{1:t}^{j}, \mathbf{w}_t^{j}\right\}_{j=1}^{N_x}$.

As time evolves, new particles are drawn from a proposal distribution, $q\left(\mathbf{x}_{1:t}|\mathbf{y}_{1:t}\right)$, which is parameterized by the sequence of states $\mathbf{x}_{1:t}$ and measurements $\mathbf{y}_{1:t}$. Similarly to \eqref{eq:filtering_bayes_}, the state at timestep $t$ is required (not the full state sequence) resulting in the weight update for particles being defined as
\begin{align}
\mathbf{w}_{t}^{j}= \mathbf{w}_{t-1}^{j}\frac{p\left(\mathbf{y}_t|\mathbf{x}_t^{j},\bm{\theta}\right)p\left(\mathbf{x}_t^{j}|\mathbf{x}^{j}_{t-1},\bm{\theta}\right)}{q\left(\mathbf{x}_t^{j}|\mathbf{x}_{t-1}^{j},\mathbf{y}_t\right)},\label{eq:weightupdate}
\end{align}
where $\mathbf{w}_{1:t-1}^{i}$\footnote{We use the notation $j$ to iterate over the particles of the PF and to differentiate between samples in the SMC sampler.} is the weight from the previous timestep.  For $t=1$,
\begin{equation}
\mathbf{w}_{1}^{j}= \frac{p\left(\mathbf{y}_1|\mathbf{x}_1^{j},\bm{\theta}\right)p\left(\mathbf{x}_1^{j}|\bm{\theta}\right)}{q\left(\mathbf{x}_1^{j}|\mathbf{y}_1\right)}.
\end{equation}
In this paper, we consider two parameterization of the weight update. The first is a derivation of the optimal proposal (see Section~\ref{example:LGSSM}) and setting the transmission model to equal the proposal (see Section~\ref{example:SIR}) such that
        \begin{align}
	q\left(\mathbf{x}_t^{j}|\mathbf{x}_{t-1}^{j},\mathbf{y}_t\right)=p\left(\mathbf{x}_t^{j}|\mathbf{x}^{j}_{t-1},\bm{\theta}\right),
	\label{prior_proposal}
	\end{align}
	which simplifies the weight update in \eqref{eq:weightupdate} as
	\begin{align}
	\mathbf{w}_{t}^{j}=\mathbf{w}_{t-1}^{j} p\left(\mathbf{y}_t|\mathbf{x}_t^{j},\bm{\theta}\right).
	\label{eq:priorproposalweight}
	\end{align}

An estimate with respect to the joint distribution, $p\left(\mathbf{y}_{1:t},\mathbf{x}_{1:t}|\bm{\theta}\right)$, can be made using the weight update in \eqref{eq:priorproposalweight}:
\begin{equation}
\int p\left(\mathbf{y}_{1:t},\mathbf{x}_{1:t}|\bm{\theta}\right)f\left(\mathbf{x}_{1:t}\right) d\mathbf{x}_{1:t} \approx \frac{1}{N_x}\sum\nolimits_{j=1}^{N_x} \mathbf{w}_{t}^{j}f\left(\mathbf{x}_{1:t}^{j}\right),\label{eq:jointexp}
\end{equation}
where $f\left(\mathbf{x}_{1:t}^{j}\right)$ denotes the function of interest on $\mathbf{x}_{1:t}^{j}$. 

A (biased)\footnote{\eqref{eq:posterior_pf} is biased
as it is a ratio of estimates.} estimate with respect to the posterior, $p\left(\mathbf{x}_{1:t}|\mathbf{y}_{1:t},\bm{\theta}\right)$ can be made by 
\begin{align}
\int p\left(\mathbf{x}_{1:t}|\mathbf{y}_{1:t},\bm{\theta}\right)&f\left(\mathbf{x}_{1:t}\right) d\mathbf{x}_{1:t} \nonumber \\
= &\int \frac{p\left(\mathbf{y}_{1:t},\mathbf{x}_{1:t}|\bm{\theta}\right)}{p\left(\mathbf{y}_{1:t}|\bm{\theta}\right)}f\left(\mathbf{x}_{1:t}\right) d\mathbf{x}_{1:t}, \label{eq:posterior_pf}
\end{align}
where 
\begin{equation}
p\left(\mathbf{y}_{1:t}|\bm{\theta}\right) = \int  p\left(\mathbf{y}_{1:t},\mathbf{x}_{1:t}|\bm{\theta}\right)d\mathbf{x}_{1:t}\approx  \frac{1}{N_x}\sum\nolimits_{j=1}^{N_x} \mathbf{w}_{t}^{j}.\label{eq:likelihood_pf_}
\end{equation}
This is in line with the joint distribution in \eqref{eq:jointexp} such that 
\begin{align}
\int  p&\left(\mathbf{x}_{1:t}|\mathbf{y}_{1:t},\bm{\theta}\right)f\left(\mathbf{x}_{1:t}\right)d\mathbf{x}_{1:t} \nonumber \\
& \approx \frac{1}{\frac{1}{N_x}\sum\nolimits_{j=1}^{N_x} {\mathbf{w}}_{t}^{j}}
\frac{1}{N_x}\sum\nolimits_{j=1}^{N_x} {\mathbf{w}}_{t}^{j}f\left(\mathbf{x}_{1:t}^{j}\right)  \\ 
&= \sum\nolimits_{j=1}^{N_x} \tilde{\mathbf{w}}_{t}^{j}f\left(\mathbf{x}_{1:t}^{j}\right),
\label{eq:a}
\end{align}
where
\begin{equation}
\tilde{\mathbf{w}}_{t}^{j} = \frac{\mathbf{w}_{t}^{j}}{\sum\nolimits_{j=1}^{N_x} \mathbf{w}_{t}^{j}}, \ \ \forall i.
\label{eq:normalisedweights}
\end{equation}
are the normalized weights. 

\subsection{Common Random Numbers}\label{sec:CRN}

The \textit{reparameterization trick} is a technique commonly used in probabilistic modeling and variational inference to enable the training of models with stochastic layers using gradient-based optimization algorithms \cite{reparam}. The \textit{reparameterization trick} expresses the randomness of a stochastic variable in a differentiable manner. This is done by reparameterizing the random variable so that its parameters are deterministic and differentiable w.r.t $\bm{\theta}$. By doing so, gradients can flow through the stochastic node during backpropagation, enabling efficient optimization. In the context of PFs, the {\itshape reparameterization trick} for resampling \cite{lee2008towards} involves fixing the random number seed in each simulation to produce CRN. The stochasticity introduced by the sampling and resampling processes are then differentiable w.r.t $\bm{\theta}$. 

\subsection{Differentiable Resampling with Common Random Numbers}\label{sec:diff_resampling}
 
As time evolves, resampling can be employed to stop particle degeneracy, where a small subset of the particles share the majority of the weight.  Monitoring the number of effective particles, $N_x^\text{eff}$, given by 
\begin{align} \label{eq: ess}
N_x^\text{eff} = \frac{1}{\sum\nolimits_{j=1}^{{N_x}} \left(\tilde{\mathbf{w}}_{t}^{j}\right)^2},
\end{align}
A common criterion for resampling is that $N_x^\text{eff}$ falls below ${{N_x}}/2$. Multiple resampling algorithms exist \cite{resamplingmethods} but within this paper we employ multinomial resampling in the PF. For estimates of gradients to be consistent before and after resampling, we implement resampling that utilizes CRN. A brief description is provided in Appendix~\ref{app:diff_resamp}, but for a more comprehensive overview we direct the reader to Section V of \cite{rosato2022efficient}.

\subsection{Log-Likelihood}\label{sec:pf_likelihood}

The log-posterior is defined as follows:
        \begin{equation}
        \log \pi(\mathbf{\bm{\theta}})=\log p(\bm{\theta})+ \log p(\mathbf{y}_{1:T}|\bm{\theta}),\label{eq:log_posterior}
        \end{equation}
        where $\log p(\bm{\theta})$ represents the log-prior and $\log p(\mathbf{y}_{1:T}|\bm{\theta})$ represents the log-likelihood.

For each time point $t=1,..,T$, an unbiased estimate of the log-likelihood\footnote{It is usually advantageous to work with log values for numerical stability.} is recursively computed by summing the log-weights as in \eqref{eq:likelihood_pf_}:
\begin{equation}
\log p\left(\mathbf{y}_{1:T}|\bm{\theta}\right)=\frac{1}{N_x}\sum_{j=1}^N \log \mathbf{w}_{t}^{j}.\label{eq:likelihood_pf}
\end{equation}

\subsection{First-Order Gradients}\label{sec:pf_gradient}

In accordance with \eqref{eq:log_posterior}, the gradient of the log-posterior is expressed as:
\begin{align}
\nabla \log \pi(\mathbf{\bm{\theta}})=\nabla \log p(\mathbf{\bm{\theta}})+\nabla \log p(\mathbf{y}_{1:T}|\mathbf{\bm{\theta}})
\label{eq:first_gradientlogposterior}
\end{align}
where $\nabla \log p(\mathbf{\bm{\theta}})$ signifies the gradient of the log-prior, and $\nabla \log p(\mathbf{y}_{1:T}|\mathbf{\bm{\theta}})$ denotes the gradient of the log-likelihood. The derivation of $\nabla \log p(\mathbf{y}_{1:T}|\mathbf{\bm{\theta}})$ is provided in Appendix~\ref{app:pf_gradient}.

\mycomment{
\subsection{Second-Order Gradients}\label{sec:second_pf_gradient}

In accordance with \eqref{eq:log_posterior}, the negative Hessian of the log-posterior is expressed as:
\begin{align}
\nabla^2 \log \pi(\mathbf{\bm{\theta}})=-\nabla^2 \log p(\mathbf{\bm{\theta}})-\nabla^2 \log p(\mathbf{y}_{1:T}|\mathbf{\bm{\theta}}),
\label{eq:second_gradientlogposterior}
\end{align}
where $\nabla^2 \log p(\mathbf{\bm{\theta}})$ signifies the gradient of the log-prior, and $\nabla^2 \log p(\mathbf{y}_{1:T}|\mathbf{\bm{\theta}})$ denotes the gradient of the log-likelihood. The derivation of $\nabla^2 \log p(\mathbf{y}_{1:T}|\mathbf{\bm{\theta}})$ is provided in Appendix~\ref{app:second_pf_gradient}.
}



\section{SMC$^2$} \label{sec:SMC_sampler}

    SMC$^2$ runs for K iterations, targeting $\pi(\mathbf{\bm{\theta}})$ ($\mathbf{\bm{\theta}} \in \mathbb{R}^{D}$)  at each iteration $k$, via IS and resampling steps which are outlined in Sections~\ref{sec:importance_sampling} and \ref{sec:smcsampler_resampling}, respectively. The joint distribution from all states until $k=K$ is defined to be
    \begin{equation}
        \pi(\mathbf{\bm{\theta}}_{1:K}) = \pi(\mathbf{\bm{\theta}}_{K}) \prod_{k=2}^{K} L(\mathbf{\bm{\theta}}_{k-1} | \mathbf{\bm{\theta}}_{k}),
    \end{equation}
    where $L(\mathbf{\bm{\theta}}_{k-1} | \mathbf{\bm{\theta}}_{k})$ is the L-kernel, which is a user-defined probability distribution. The choice of this distribution can affect the efficiency of the sampler \cite{green2022increasing}.

    \subsection{Importance Sampling}\label{sec:importance_sampling}
        At $k=1$, $N$ samples $\forall i = 1, \dots, N$ are drawn from a prior distribution $q_1(\cdot)$ as follows:
        \begin{equation} \label{eq: init_sample}
            \mathbf{\bm{\theta}}^i_1 \sim q_1(\cdot), \ \ \forall i,
        \end{equation}
        and weighted according to
        \begin{equation} 
            \mathbf{w}^i_1 = \frac{\pi(\mathbf{\bm{\theta}}^i_1)}{q_1(\mathbf{\bm{\theta}}^i_1)}, \ \ \forall i\label{init_weights}.
        \end{equation}
        
        At $k>1$, subsequent samples are proposed based on samples from the previous iteration via a proposal distribution, $q(\mathbf{\bm{\theta}}^i_k|\mathbf{\bm{\theta}}^i_{k-1})$, as follows: 
        \begin{equation} \label{eq: proposal}
            \mathbf{\bm{\theta}}^i_k \sim q(\cdot|\mathbf{\bm{\theta}}^i_{k-1}).
        \end{equation}
        These samples are weighted according to 
        \begin{equation}
            \mathbf{w}^i_{k} = \mathbf{w}^i_{k-1} \frac{\pi(\mathbf{\bm{\theta}}^i_{k})}{\pi(\mathbf{\bm{\theta}}^i_{k-1})} \frac{L(\mathbf{\bm{\theta}}^i_{k-1}|\mathbf{\bm{\theta}}^i_{k})}{q(\mathbf{\bm{\theta}}^i_{k}|\mathbf{\bm{\theta}}^i_{k-1})}, \ \ \forall i.\label{eq:l_weights}
        \end{equation}
        Estimates of the expectations of functions, such as moments, on the distribution are realised by 
        \begin{equation} 
            \Tilde{\mathbf{f}}_{k} = \sum\nolimits_{i=1}^{N} \Tilde{\mathbf{w}}^i_{k} \mathbf{\bm{\theta}}^i_{1:k}, \label{realised_estimates}
        \end{equation}
        where the normalized weights $\Tilde{\mathbf{w}}^i_{k}$ are calculated as in \eqref{eq:normalisedweights}:
        \begin{equation}
            \tilde{\mathbf{w}}_{k}^{i} = \frac{\mathbf{w}_{k}^{i}}{\sum\nolimits_{j=1}^{N} \mathbf{w}_{k}^{j}}, \ \ \forall i.
            \label{eq: normalise_smc2}
        \end{equation} 
        Recycling constants are also calculated (see Section \ref{sec:recycling}):
        \begin{equation}
                \mathbf{c}_k = \frac{\mathbf{l}_k}{\sum\nolimits_{k=1}^K \mathbf{l}_k}, 
                \ \ \forall k,\label{eq: opt_recycling_constant}
            \end{equation}
            where 
            \begin{align} \label{eq: recycling_I}
                \mathbf{l}_k = \frac{\left(\sum\nolimits_{i=1}^{{N}} 
                \mathbf{w}_{k}^{i}\right)^2}{\sum\nolimits_{i=1}^{{N}} 
                \left(\mathbf{w}_{k}^{i}\right)^2}.
                \ \ \forall k.
            \end{align}
        Sample degeneracy occurs for the same reasoning as the PF and is mitigated by resampling.
        
        \subsection{$\mathcal{O}(\log_2N)$ Resampling in the SMC Sampler}\label{sec:smcsampler_resampling}
        The SMC sampler layer of SMC$^2$ computes the Effective Sample Size (ESS) in the same way as the PF in \eqref{eq: ess}: 
            \begin{align} \label{eq: ess_smc2}
                N^\text{eff} = \frac{1}{\sum\nolimits_{i=1}^{{N}} \left(\tilde{\mathbf{w}}_{k}^{i}\right)^2}.
            \end{align}
            Resampling is performed at iteration $k$ if $N^\text{eff} < N/2$.
        
        Resampling is widely known to be a bottleneck when attempting to parallelize SMC samplers \cite{Alessandro, Alessandro2, Alessandro5}. In this work, the fully parallelized resampling algorithm outlined and implemented in \cite{Alessandro5, rosato2023mathcal, drousiotis2023shared} is utilized. With $P$ computer cores, the parallelization achieves $\mathcal{O}(\frac{N}{P} + \log_2P) \rightarrow \mathcal{O}(\log_2N)$ time complexity which is proven to be optimal. We do not describe parallel resampling in this paper, but refer the reader to \cite{Alessandro, rosato2023mathcal} for a comprehensive description including proofs. 
    
        \subsection{Recycling}\label{sec:recycling}
            In using \eqref{realised_estimates} to calculate estimates, only samples from the most recent iteration are used. Estimates from previous iterations can be utilized through
            \begin{equation}
                \Tilde{\mathbf{f}} = \sum\nolimits_{k=1}^K \mathbf{c}_k\Tilde{\mathbf{f}}_k,
            \end{equation}
            where $\sum\nolimits_{k=1}^K \mathbf{c}_k = 1$. The optimal recycling constants in \eqref{eq: opt_recycling_constant} maximize the ESS of the whole population of samples.

        \subsection{Proposals}\label{sec:proposals}

        In this paper, the two proposals considered include the RW and the Langevin proposal that includes first-order gradients which are denoted SMC0 and SMC1, respectively. They are defined to be:
        \[
            q(\bm{\theta}^i_k|\bm{\theta}^i_{k-1})= 
        \begin{cases}
            \mathcal{N}(\bm{\theta}^i_k; \bm{\theta}^i_{k-1}, \bm{\Gamma}),& \text{[SMC0]}\\
            \mathcal{N}(\bm{\theta}^i_k; \bm{\theta}^i_{k-1} +  F(\bm{\theta}^i_{k-1}), \bm{\Gamma}),              & \text{[SMC1]}.
        \end{cases}
\] where
        \begin{itemize}
              \item $\bm{\Gamma}=\gamma^{2} \mathbf{I}_{d}$, for step size $\gamma$,
              \item $E(\mathbf{\bm{\theta}}^i_{k-1}) = \nabla \log \pi(\mathbf{\bm{\theta}}^i_{k-1})$,
              \item $F(\mathbf{\bm{\theta}}^i_{k-1}) = \frac{1}{2} \bm{\Gamma} E(\mathbf{\bm{\theta}}^i_{k-1})$,
        \end{itemize}
        The introduction of the gradient of the log-posterior \eqref{eq:first_gradientlogposterior} into the proposal has been shown to improve the rate of convergence empirically and analytically under certain conditions \cite{Second-order, inproceedings}.   
        \begin{align}
            \begin{split}                      q(\mathbf{\bm{\theta}}^i_k|\mathbf{\bm{\theta}}^i_{k-1})= 
            \mathbf{\bm{\theta}}^i_{k-1} + \mathbf{\Gamma} &\nabla \log p(\mathbf{\bm{\theta}}^i_{k-1}|\mathbf{y}_{1:T}) \\
            + &\mathcal{N}(\mathbf{p}^i_{k-1};0, \mathbf{\Gamma}), \ \ \forall i.
            \end{split}
        \end{align}
    
        \subsection{L-kernels}\label{sec:l_kernels}
            Langevin is effectively equivalent to HMC with only a single timestep, so the dynamics can be written as  
            \begin{align}
              \mathbf{p}^i_k = \mathbf{p}^i_{k-1} + F(\mathbf{\bm{\theta}}^i_{k-1}), \nonumber\\
              \bm{\theta}^i_k = \bm{\theta}^i_{k-1} + \Gamma \mathbf{p}^i_k.
            \end{align}
            
            Using the logic of \cite{devlin2021no}; with $f_{LMC}$ taking the place of $f_{LF}$ as the function transforming parameter $\bm{\theta}^i_{k-1}$ to $\bm{\theta}^i_k$ (i.e. 
            $\bm{\theta}^i_k$=$f_{LMC}      (\bm{\theta}^i_{k-1}, \mathbf{p}^i_{k-1})$); the proposal can be written
        \begin{align}
        \begin{split}
               q(&\bm{\Theta}_k=f_{LMC}      (\bm{\theta}^i_{k-1}, \mathbf{p}^i_{k-   1})|\bm{\Theta}_{k-       1}=\bm{\theta}^i_{k-1}) \\
                    & = q(\mathbf{P}_{k-1}=\mathbf{p}^i_{k-     1}|\bm{\Theta}_{k-1}=\bm{\theta}^i_{k-      1})\begin{vmatrix}
                \frac{df_{LMC}(\bm{\theta}^i_{k-1},         \mathbf{p}^i_{k-1})}{d\mathbf{p}^i_{k-      1}}
            \end{vmatrix}^{-1} \\
            & = \mathcal{N}(\mathbf{p}^i_{k-1};         0,\bm{\Gamma}) \begin{vmatrix}
                \frac{df_{LMC}(\bm{\theta}^i_{k-1},         \mathbf{p}^i_{k-1})}{d\mathbf{p}^i_{k-      1}}
            \end{vmatrix}^{-1}, \ \ \forall i,
        \end{split}
        \end{align}      
        and the L-kernel as        
        \begin{align}
            \begin{split}
                 L(\bm{\Theta}_{k-1}=&f_{LMC}         (\bm{\theta}^i_k, -      \mathbf{p}^i_k) |\bm{\Theta}_k=\bm{\theta}^i_            k) \\
                & = L(\mathbf{P}_k=-            \mathbf{p}^i_k|\bm{\Theta}_k=\bm{\theta}^i_k            )\begin{vmatrix}
                    \frac{df_{LMC}(\bm{\theta}^i_k, -           \mathbf{p}^i_k)}{d\mathbf{p}^i_k}
                \end{vmatrix}^{-1} \\
                & = \mathcal{N}(-\mathbf{p}^i_k;            0,\bm{\Gamma}) \begin{vmatrix}
                    \frac{df_{LMC}(\bm{\theta}^i_k, -           \mathbf{p}^i_k)}{d\mathbf{p}^i_k}
                \end{vmatrix}^{-1}, \ \ \forall i.\label{eq: l=q}
            \end{split}
        \end{align}

\section{Numerical Results}\label{sec:Examples}

\begin{figure}[h]
    \centering
    \begin{subfigure}[h]{0.24\textwidth}
        \includegraphics[width=\linewidth]{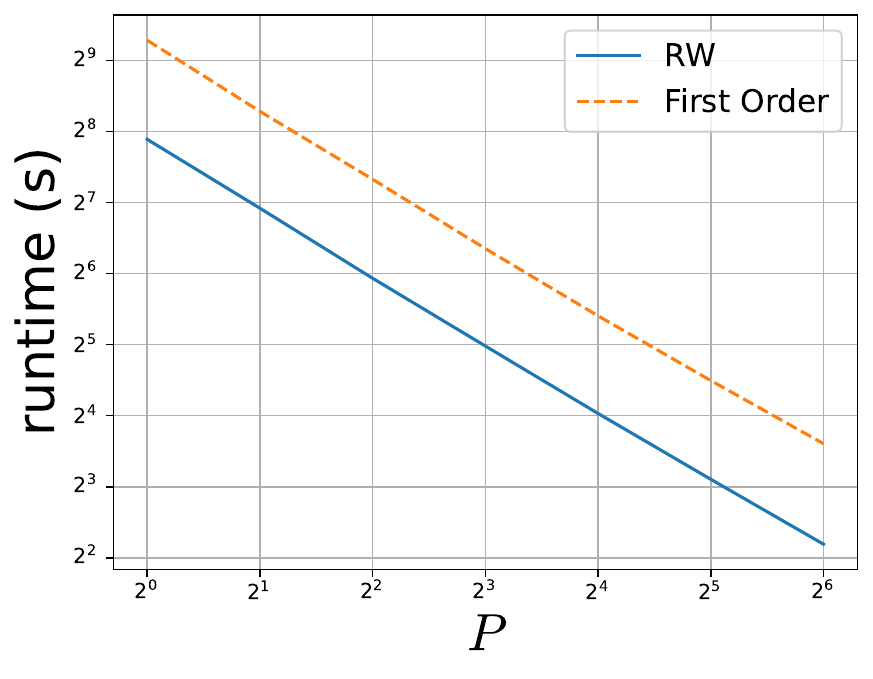}
        \caption{} \label{fig:runtimes}
    \end{subfigure}
    \begin{subfigure}[h]{0.24\textwidth}
        \includegraphics[width=\linewidth]{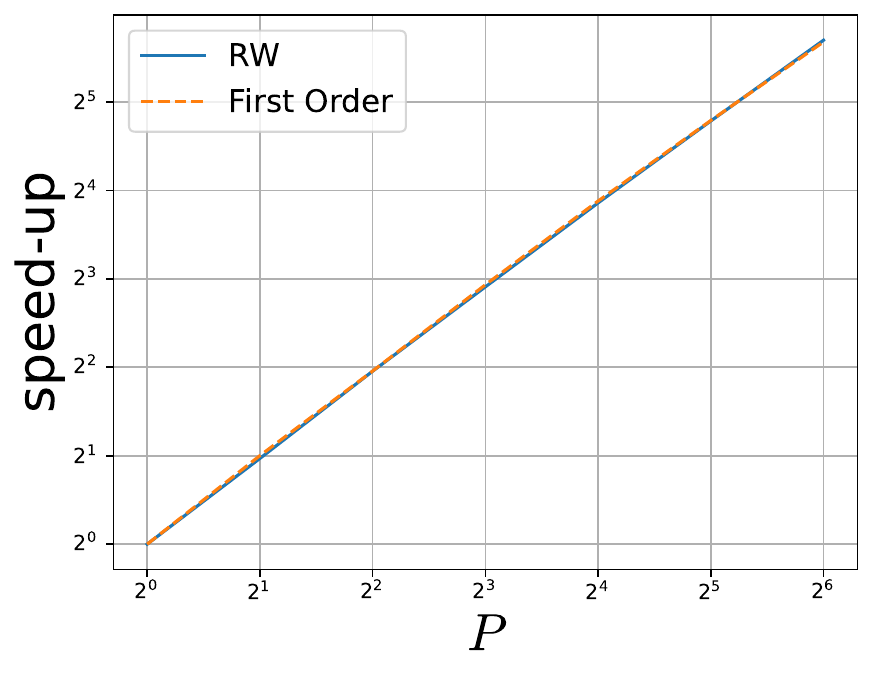}
        \caption{} \label{fig:_speedup}
    \end{subfigure}
        
    \caption{(a) Runtime and (b) speed-up plots of the LGSSM when changing number of $P = 1, 2, 4, \dots, 64$ computational cores.}
    \label{fig:speedip_runtime}
\end{figure}

In this paper, we use the distributed SMC$^2$ framework described in \cite{rosato2023mathcal}. The time complexities of each process are outlined in Table~\ref{tab: smc2_tasks}. The analysis conducted in this paper was run on a cluster of $2$ distributed memory (DM) machines. Each DM machine consists of a `2 Xeon Gold 6138' CPU, which provides a memory of $384$GB and $40$ cores. Figure~\ref{fig:speedip_runtime}(a) and (b) shows the runtime and speed-up, respectively, of RW and first-order proposals for the LGSSM outlined in Section~\ref{example:LGSSM}. The analysis was run with $P = 1, 2, 4, \dots, 64$ cores. This is because the parallelization strategy for the $\mathcal{O}(\log_2N)$ resampling algorithm, outlined in Section~\ref{sec:smcsampler_resampling}, utilizes the divide-and-conquer paradigm. Having a power-of-two number for $P$ optimizes the workload balance across the processors. For the same reasoning as choosing $P$, the number of $N$ samples within the SMC sampler is also chosen to be a power-of-two to balance the workload across the $P$ computational cores. The results presented in Figure~\ref{fig:LGSSM_convergence} are averaged over 5 Monte Carlo runs. The source code is available here: \url{https://github.com/j-j-murphy/SMC-Squared-Langevin}.

    \begin{table}[] 
    \begin{center}
        \centering
        \small
        \caption{Time complexity of each process of SMC$^2$ on DM using $N$ samples and $P$ computational cores.\label{tab: smc2_tasks}}
        \begin{tabular}{cc}
            \toprule
            \toprule
            \textbf{\shortstack{Task name - Details}} & \textbf{\shortstack{Time complexity}}\\
            \midrule
            IS - Eq. \eqref{eq: proposal} \& \eqref{eq:l_weights}		& $\mathcal{O}(\frac{N}{P}) \rightarrow \mathcal{O}(1)$\\
            First-order gradient - Eq. \eqref{eq:first_gradientlogposterior}			& $\mathcal{O}(\frac{N}{P}) \rightarrow \mathcal{O}(1)$\\
            Normalise - Eq. \eqref{eq: normalise_smc2}			& $\mathcal{O}(\frac{N}{P} + \log_2P) \rightarrow \mathcal{O}(\log_2N)$\\
            ESS - Eq. \eqref{eq: ess_smc2}		& $\mathcal{O}(\frac{N}{P} + \log_2P) \rightarrow \mathcal{O}(\log_2N)$\\
            Estimate - Eq. \eqref{realised_estimates}			& $\mathcal{O}(\frac{N}{P} + \log_2P) \rightarrow \mathcal{O}(\log_2N)$\\
            Recycling - Eq. \eqref{eq: recycling_I}			& $\mathcal{O}(\frac{N}{P} + \log_2P) \rightarrow \mathcal{O}(\log_2N)$\\
            \bottomrule
            \bottomrule
        \end{tabular}
    \end{center}
\end{table}

\subsection{Linear Gaussian State Space Model}\label{example:LGSSM}

We consider the Linear Gaussian State Space (LGSS) model outlined in \cite{dahlin2019getting, rosato2022efficient} which is given by
\begin{equation}
\mathbf{x}_{t} \mid \mathbf{x}_{t-1} \sim \mathcal{N}\left(\mathbf{x}_{t} ; \mu \mathbf{x}_{t-1}, \phi^{2}\right),
\label{LGSSXt}
\end{equation}
\begin{equation}
\mathbf{y}_{t} \mid \mathbf{x}_{t} \sim \mathcal{N}\left(\mathbf{y}_{t} ; \mathbf{x}_{t}, \sigma^{2}\right),
\label{LGSSYt}
\end{equation}
where $\bm{\theta}=\left\{\mu, \phi, \sigma \right\}$ with Uniform prior densities $\mathcal{U}(-1,1)$, $\mathcal{U}(0,5)$ and $\mathcal{U}(0,5)$, respectively. The ``optimal'' proposal is used within the PF and can be derived from the properties of \eqref{LGSSXt} and \eqref{LGSSYt}:
\begin{equation}
q\left(\mathbf{x}_{t}|\mathbf{x}_{t-1},\mathbf{y}_t\right) =\mathcal{N}\left(\mathbf{x}_{t} ; \rho^{2}\left[\sigma^{-2} \mathbf{y}_{t}+\phi^{-2} \mu \mathbf{x}_{t-1}\right], \rho^{2}\right),
\end{equation}
with $\rho^{-2}=\phi^{-2}+\sigma^{-2}$. The PF weights are updated using
\begin{equation}
\mathbf{w}_{1:t}^{j}=\mathbf{w}_{1:t-1}^{j}\mathcal{N}\left(\mathbf{y} ; \mu \mathbf{x}_{t}, \rho^2\right).
\end{equation}
The experimental configuration comprises $T=500$ observations, with $N_x=250$ particles in the PF and $N=64$ particles in the SMC sampler. The setup involves $K=15$ iterations. As step-size adaption was not implemented in this paper, the step-sizes were hard-coded before the simulation started. To determine an appropriate step size, multiple runs were conducted across a range of step sizes, and the one yielding the best performance was selected. The step-sizes for the RW and first-order proposal were fine-tuned to 0.175 and 0.085, respectively.

\begin{figure*}

            \centering
            \begin{subfigure}[t]{0.32\textwidth}
                \includegraphics[width=\linewidth]{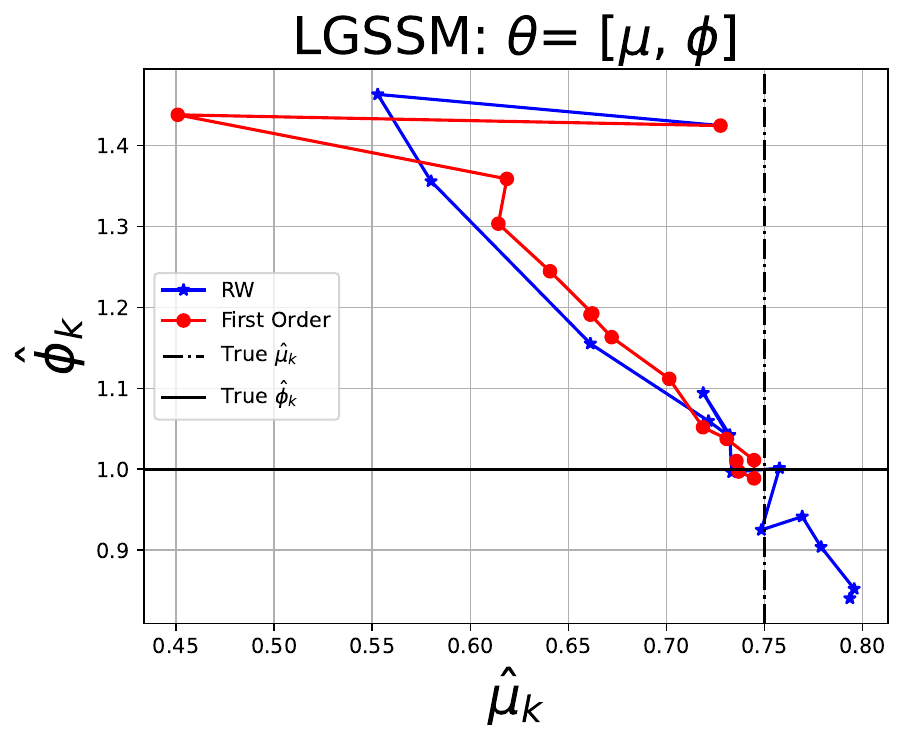}
                \caption{$\bm{\theta}=\left[\mu, \phi\right]$} \label{fig: stm_hard_fl_smc}
            \end{subfigure}
            \begin{subfigure}[t]{0.32\textwidth}
                \includegraphics[width=\linewidth]{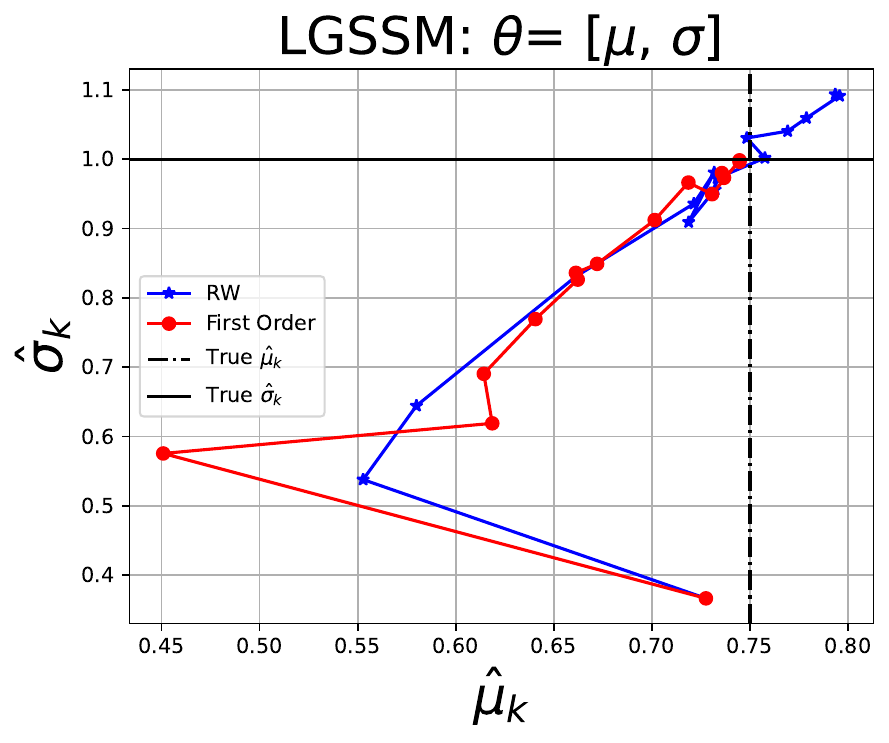}
                \caption{$\bm{\theta}=\left[\mu, \sigma\right]$} \label{fig: stm_hard_fl_ekf1}
            \end{subfigure}
            \begin{subfigure}[t]{0.32\textwidth}
                \includegraphics[width=\linewidth]{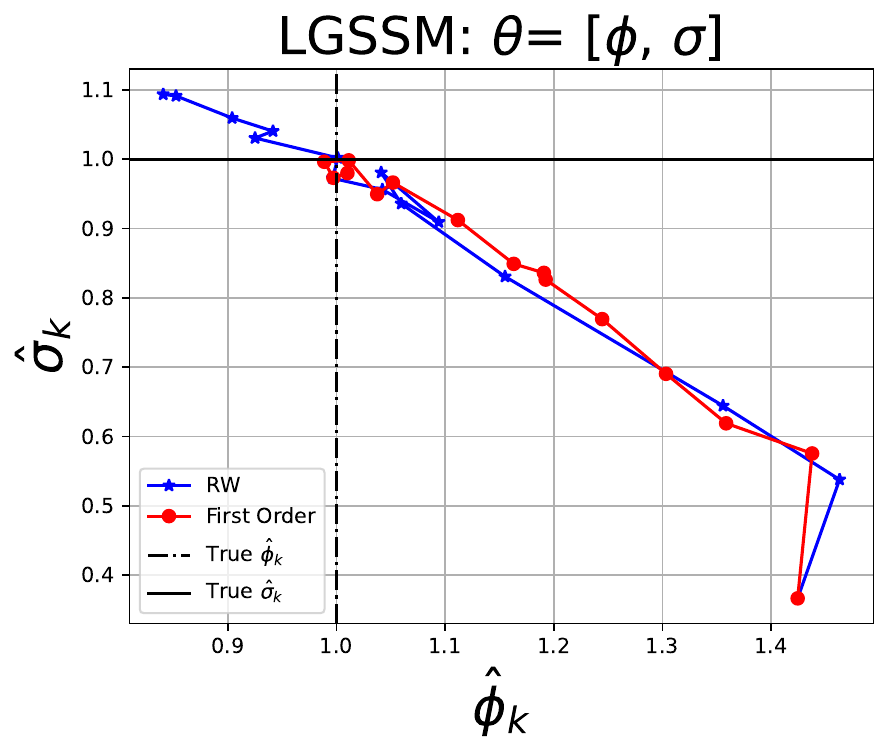}
                \caption{$\bm{\theta}=\left[\phi, \sigma\right]$} \label{fig: stm_hard_fl_ekf}
            \end{subfigure}
        
            \caption{Convergence plots depicting the parameters of the LGSSM in two dimensions when using the RW (blue line with star markers) and first-order (red line with circle markers) proposals over $K=15$ iterations and average over 5 Monte-carlo runs. The true values are outlined with the black solid and dot/dashed lines.}
            \label{fig:LGSSM_convergence}
        \end{figure*}

Incorporating first-order gradients in the proposal yields more precise parameter estimates when compared to the RW method, as seen in Table~\ref{Table:LGSSM_results}. The first-order proposal also performs better in terms of the effective sample size. Visual representations of convergence are provided in Figure~\ref{fig:LGSSM_convergence}. It is apparent that leveraging gradient information enables the model to approach and maintain proximity to the true parameters, as indicated by the dashed vertical and horizontal blue lines. 

\begin{table}[]
\caption{LGSSM: Parameter estimation results for the LGSSM when employing RW and first-order proposals. The true values are: $\mu = 0.75$, $\phi = 1$ and $\sigma = 1$.}
\renewcommand{\arraystretch}{1.2}
\centering
\begin{tabular}{cccccc}
\hline 
\hline 
\textbf{Proposal} & $\mathbf{E[\mu]}$ & $\mathbf{E[\phi]}$ & $\mathbf{E[\sigma]}$ & \textbf{MSE} & \textbf{ESS}\\
\hline 
RW & $0.793$ & $0.841$ & $1.093$ & $0.088$ & $0.052$\\
First-order & $0.736$ & $1.010$ & $0.980$ & $2.3 \times 10^{-4}$ & $0.106$\\
\hline \hline
\end{tabular}
\label{Table:LGSSM_results}
\end{table}

\subsection{Susceptible, Infected and Recovered Disease Model}\label{example:SIR}

In this example we consider the Susceptible, Infected and Recovered (SIR) epidemiological disease model outlined in \cite{sheinson2014comparison, rosato2023disease, rosato2022inference}. A discrete time approximation of the SIR model is presented below
\begin{align}
&\mathbf{S}_{t}=\mathbf{S}_{t-1}-\beta \mathbf{I}_{t-1} \mathbf{S}_{t-1}+\epsilon_{\beta}, \label{SIRR:s}\\
& \mathbf{I}_{t}= \mathbf{I}_{t-1}+\beta  \mathbf{I}_{t-1} \mathbf{S}_{t-1}-\gamma \label{SIRR:i}  \mathbf{I}_{t-1}-\epsilon_{\beta}+\epsilon_{\gamma},\\
& \mathbf{R}_{t}=N_{pop} - \mathbf{S}_{t} - \mathbf{I}_{t} \label{SIRR:r},
\end{align}
where $\beta$ is the mean rate of people an infected person infects per day and $\gamma$ is the proportion of infected recovering per day which we define as $\bm{\theta}=\left[\beta, \gamma\right]$ with Uniform prior densities $\mathcal{U}(0,1)$ and $\mathcal{U}(0,1)$. The total amount of individuals in the susceptible, infected and recovered compartments at time $t=0$ are denoted $\mathbf{S}_{0}=N_{pop}-\mathbf{I}_{0}$, $\mathbf{I}_{0}=1$ and $\mathbf{R}_{0}=0$, respectively. The total population within the simulation is denoted $N_{pop}=763$. Stochasticity is introduced when modelling the dynamics by including a noise term, $\epsilon_x$, for each time-varying parameter, $x$, which mimics the interactions between people in the population, $N_{pop}$. Note $\epsilon_x$ will be independent for different values of $x$ and are drawn from $\epsilon_{x} \sim \mathcal{N}(0, 0.5)$.

In this example, we set the transmission model to equal the proposal as in \eqref{prior_proposal}.

The experimental configuration comprises $T=35$ observations, with $N_x=250$ particles in the PF and $N=64$ particles in the SMC sampler. The setup involves $K=15$ iterations. The step-sizes for the RW and first-order proposal were fine-tuned to 0.01 and 0.006, respectively.

\begin{figure*}
    \centering
            \begin{subfigure}[t]{0.32\textwidth}
                \includegraphics[width=\linewidth]{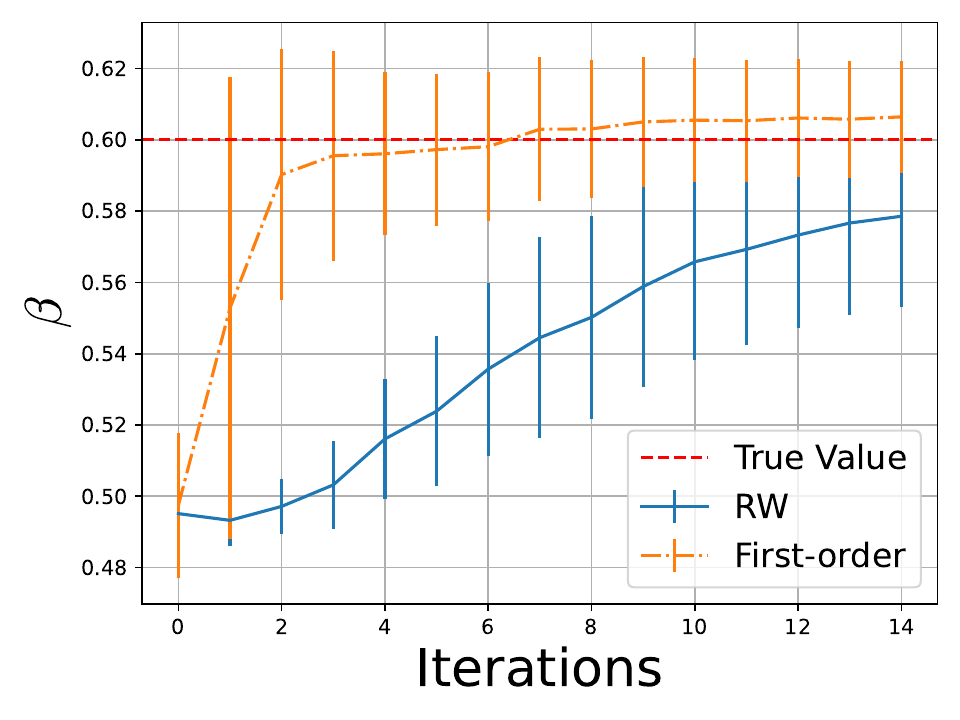}
                \caption{} \label{fig:1}
            \end{subfigure}
            \begin{subfigure}[t]{0.32\textwidth}
                \includegraphics[width=\linewidth]{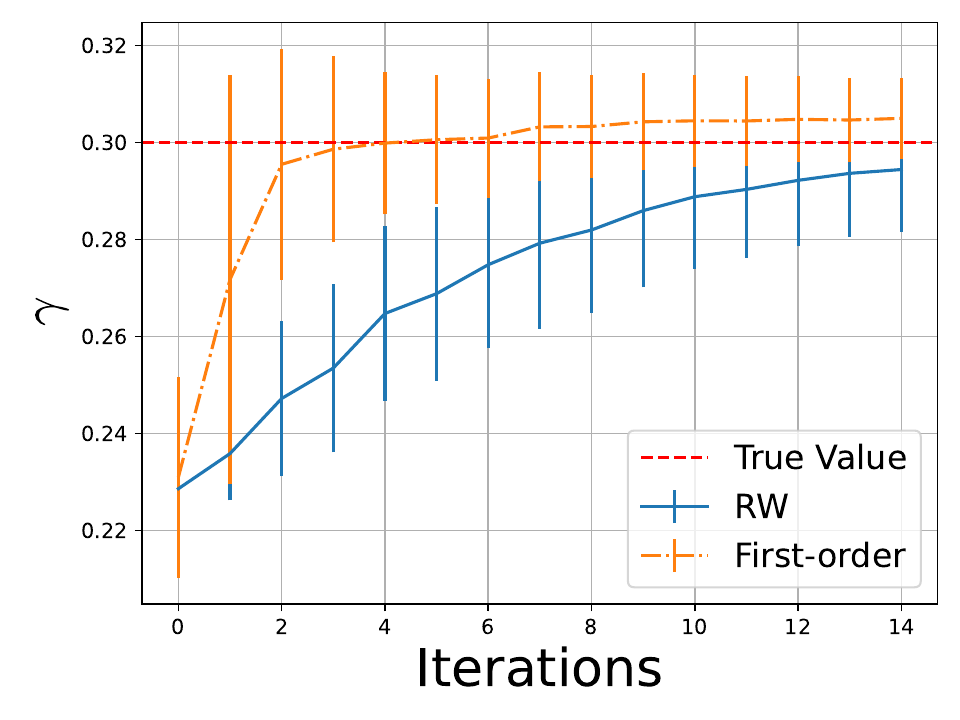}
                \caption{} \label{fig:2}
            \end{subfigure}
            \begin{subfigure}[t]{0.32\textwidth}
                \includegraphics[width=\linewidth]{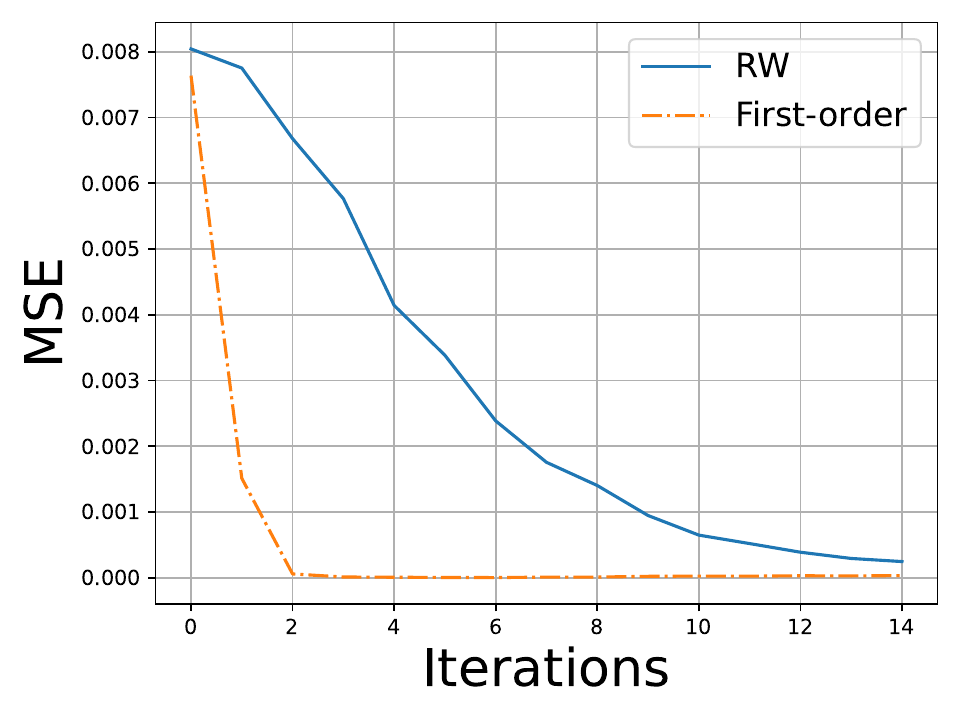}
                \caption{} \label{fig:3}
            \end{subfigure}
            \caption{Convergence plots of (a) $\beta$ and (b) $\gamma$ of the SIR disease model outlined in \eqref{SIRR:s}-\eqref{SIRR:r} when using the RW (blue solid line) and first-order (orange dot/dashed line) proposals over $K=15$ iterations and averaged over 5 Monte-carlo runs. The true values are signalled by the horizontal red dashed line. The average MSE of $\beta$ and $\gamma$ plotted at each iteration can be seen in (c).}
            \label{fig:SIR_convergence}
        \end{figure*}

Similarly to the LGSSM, Table~\ref{Table:SIR_result} outlines that using first-order gradients within the proposal outperforms the RW proposal in terms of accuracy and the ESS when estimating the parameters of the SIR disease model. The convergence plots can be seen visually in subplots (a) and (b) in Figure~\ref{fig:SIR_convergence}. Subplot (c) shows the rolling average MSE of both parameters over the 15 iterations. It is evident that the MSE drops to 0 in a quicker time-frame when using the first-order when compared to the RW.

\begin{table}[]
\caption{SIR: Parameter estimation results for the SIR disease model when employing RW and first-order proposals. The true values are: $\beta = 0.6$ and $\gamma = 0.3$.}
\renewcommand{\arraystretch}{1.2}
\centering
\begin{tabular}{ccccc}
\hline 
\hline 
\textbf{Proposal} & $\mathbf{E[\beta]}$ & $\mathbf{E[\gamma]}$& \textbf{MSE} & \textbf{ESS}\\
\hline 
RW & 0.581 & 0.295 & $2.01 \times 10^{-4}$ &0.104\\
First-order& 0.604  & 0.304 & $1.69 \times 10^{-5}$ &0.241\\
\hline \hline
\end{tabular}
\label{Table:SIR_result}
\end{table}

\section{Conclusions and Future Work}\label{sec:conclusions}

This paper extends the SMC$^2$ framework introduced in \cite{rosato2023mathcal} by integrating first-order gradients estimated from a CRN-PF into a Langevin proposal. We show that utilising first-order gradients can provide more accurate parameter estimates in two SSMs.

While integrating first-order gradients into the Langevin proposal is better than RW, we suspect that that using a HMC or NUTS proposal would provide similar additional benefits to that seen in \cite{rosato2022efficient} for p-MCMC. Employing gradient based proposals within the PF, as well as the SMC sampler, could be a fruitful direction for future work \cite{varsi2024general}.

It is important to acknowledge that these enhancements are anticipated to increase the runtime of SMC$^2$. However, this potential drawback could be mitigated by extending the DM parallelization approach outlined in this paper to hybrid memory architectures. For instance, leveraging clusters of DM GPUs could offer a viable solution.




\section*{Acknowledgment}

The authors would like to thank Andrew Millard for helping with the derivation of the Langevin L-kernel.

\bibliographystyle{ieeetr}
\bibliography{bibliography.bib}

\appendix

\subsection{First-Order Gradients}\label{app:pf_gradient}

Given \eqref{eq:likelihood_pf}, an approximation of the gradient of the likelihood is determined as a function of the derivative of the weights:
\begin{eqnarray}
	\frac{d}{d\bm{\theta}} {p(\mathbf{y}_{1:t}|\bm{\theta})} & \approx & \frac{1}{N} \sum_{i=1}^N \frac{d}{d\bm{\theta}}\mathbf{w}_{1:t}^{j}. \label{eq:dlikweightsum_grad}
\end{eqnarray}
Applying the chain rule, the derivative of the log-likelihood in \eqref{eq:first_gradientlogposterior} is given by:
\begin{align}
	\frac{d}{d\bm{\theta}} {\log p(\mathbf{y}_{1:t}|\bm{\theta})} &\approx \frac{1}{N} \frac{1}{p(\mathbf{y}_{1:t}|\bm{\theta})}\sum_{i=1}^N \mathbf{w}_{1:t}^{j}\frac{d}{d\bm{\theta}} {\log \mathbf{w}_{1:t}^{j}}
		\\ &\approx \frac{1}{N}
	\sum_{i=1}^N \mathbf{\tilde{w}}_{1:t}^{j}\frac{d}{d\bm{\theta}} {\log \mathbf{w}_{1:t}^{j}},	\label{eqn:dloglik}
\end{align}
where
\begin{equation}
    \mathbf{\tilde{w}}_{1:t}^{j} = \frac{\mathbf{w}_{1:t}^{j}}{p(\mathbf{y}_{1:t}|\bm{\theta})}
    \label{eq:normalised_weights}
\end{equation}
are normalized weights. Following \eqref{eq:priorproposalweight}, an estimate of the derivative of the log weights is given by:
\begin{align}	\label{eqn:weightupdate}
	\frac{d}{d\bm{\theta}} {\log \mathbf{w}_{1:t}^{j}} &= \frac{d}{d\bm{\theta}} {\log \mathbf{w}_{1:t-1}^{j}} + \frac{d}{d\bm{\theta}} {\log p\left(\mathbf{y}_t | \mathbf{x}_{t}^{j}\right)}.
\end{align}



Let 
\begin{align}
    	L\left(\mathbf{x}_t^{j}, \bm{\theta}, \mathbf{y}_t\right) &\triangleq \log p\left(\mathbf{y}_t | \mathbf{x}_{t}^{j}\right), \label{eq:L}
\end{align}
where the likelihood is Gaussian with a variance that is independent of $x^{j}_t$, such that
\begin{align}
L\left(\mathbf{x}_t^{j}, \bm{\theta}, \mathbf{y}_t\right)&\triangleq \logN\left(\mathbf{y}_t; h(x^{j}_t, \bm{\theta}), R(\bm{\theta})\right).
\end{align}
Application of the chain rule yields
\begin{align}\label{eqn:deriv_loglike}
	\frac{d}{d\bm{\theta}} L\left(\mathbf{x}_t^{j}, \bm{\theta}, \mathbf{y}_t\right) =&
		\frac{\partial}{\partial h}\logN(\mathbf{y}_t; h, R)\left(\frac{\partial h}{\partial \mathbf{x}^{j}_t}\frac{d \mathbf{x}^{j}_t}{d \bm{\theta}} +
			\frac{\partial h}{\partial \bm{\theta}}\right) \nonumber \\ &+ \frac{\partial}{\partial R}\logN(\mathbf{y}_t; h, R)\frac{d R}{d \bm{\theta}}.
\end{align}

Suppose the proposal takes the form
\begin{align}
	q(\mathbf{x}_t^{j}| &\mathbf{x}_{t-1}^{j}, \bm{\theta}, \mathbf{y}_t) = \Normal(\mathbf{x}_t^{j}; \mu(\mathbf{x}^{j}_{t-1}, \bm{\theta}, \mathbf{y}_t), C(\mathbf{x}^{j}_{t-1}, \bm{\theta}, \mathbf{y}_t)).
		\label{eqn:proposal}
\end{align}
If the proposal noise $\bm{\epsilon}^{j}_t \sim \mathcal{N}(\cdot; 0, I_\nx)$ is sampled in advance, the \textit{reparameterization trick} \cite{reparam} can be employed such that the new particle states can be written as a deterministic function:
\begin{align}
	\mathbf{x}^{j} &= f(\mathbf{x}^{j}_{t-1}, \bm{\theta}, \mathbf{y}_t,\bm{\epsilon}^{j}_t) \nonumber \\
	&\triangleq \mu(\mathbf{x}^{j}_{t-1}, \bm{\theta}, \mathbf{y}_t) + \sqrt{C(\mathbf{x}^{j}_{t-1}, \bm{\theta}, \mathbf{y}_t)}\times \bm{\epsilon}^{j}_t. \label{eq:reparam_epsilon}
\end{align}
By the chain rule, the derivative is given by
\begin{eqnarray}
	\frac{d\mathbf{x}_t^{j}}{d\bm{\theta}} & = & \frac{d}{d\bm{\theta}}f(\mathbf{x}_{t-1}^{j}, \bm{\theta}, \mathbf{y}_t, \bm{\epsilon}^{j}_t) \label{eqn:df} \\ & = &
		\frac{\partial f}{\partial \mathbf{x}_{t-1}^{j}}\frac{d \mathbf{x}_{t-1}^{j}}{d \bm{\theta}} + \frac{\partial f}{\partial\bm{\theta}} \label{eqn:partialf}
		\frac{d \bm{\theta}}{d \bm{\theta}} \\ & = & 
		\frac{\partial f}{\partial \mathbf{x}_{t-1}^{j}}\frac{d \mathbf{x}_{t-1}^{j}}{d \bm{\theta}} + \frac{\partial f}{\partial\bm{\theta}}.	\label{eqn:dxk}
\end{eqnarray}
\subsection{Differentiable Resampling using Common Random Numbers}\label{app:diff_resamp}
 Let
\begin{eqnarray}
	c^{i_t} = \frac{\sum_{j=1}^i \mathbf{w}_{1:t}^{j}}{\sum_{j=1}^N \mathbf{w}_{1:t}^{j}}
\end{eqnarray}
be the normalized cumulative weights and let 
\begin{align}
	\resample_i= \resample\left(\resampleu_t^i, \mathbf{w}_{1:t}^{1:N}\right) = \sum_{j=0}^{N-1}\left[\resampleu_t^i > c^{j}_t \right]
\end{align}
be the index sampled for particle $i$, where $\resampleu_t^i \sim \mbox{Uniform}((0, 1])$ are independent for each particle and timestep. The particle indices are sampled according to a categorical distribution, giving a multinomial resampler, where each index is resampled with a probability proportional to its weight.

The resampled weights are equal, while preserving the original sum:
\begin{eqnarray}
	\mathbf{x}^{\prime i}_t & = & \mathbf{x}_t^{\kappa_i}, \\
	\mathbf{w}^{\prime i}_{1:t} & = & \frac1N \sum_{j=1}^N \mathbf{w}_{1:t}^{j}.	\label{eqn:weightresample}
\end{eqnarray}
From \eqref{eqn:weightresample}, it is clear that
\begin{eqnarray}
	\frac{d}{d\bm{\theta}} \mathbf{}w^{\prime i}_{1:t} & = & \frac1N \sum_{j=1}^N \ddtheta \mathbf{w}_{1:t}^{j}. \label{eqn:weightdiffresample}
\end{eqnarray}
Converting this derivative to the derivative of log weights, application of the chain rule gives
\begin{align}
	\frac{d}{d\bm{\theta}} \log \mathbf{w}^{\prime i}_{1:t} &= \frac1N\frac1{\mathbf{w}^{\prime i}_{1:t}}\sum_{j=1}^N \mathbf{w}_{1:t}^{j}\ddtheta \log \mathbf{w}_{1:t}^{j}
		\\ &=
	\sum_{j=1}^N \tilde{\mathbf{w}}_{1:t}^{j}\ddtheta \log \mathbf{w}_{1:t}^{j},
\end{align}
where $\tilde{\mathbf{w}}_{1:t}^{j}$ are the normalized weights.

To obtain the particle gradient, note that, for differentiable $\resample$,
\begin{align}
	\frac{d}{d\bm{\theta}} \mathbf{x}^{\prime i}_t =&
		\frac{\partial}{\partial\resample}\mathbf{x}_t\left(\bm{\theta}, \resample\left(\resampleu_t^i, \mathbf{w}_{1:t}^{1:N}\right)\right) \frac{\partial}{\partial\bm{\theta}}\resample\left(\resampleu_t^i, \mathbf{w}_{1:t}^{1:N}\right) \\ &+ \frac{d}{d\bm{\theta}}\mathbf{x}_t\left(\bm{\theta}, \resample\left(\resampleu_t^i, \mathbf{w}_{1:t}^{1:N}\right)\right).
\end{align}
Since $\partial\resample/\partial\bm{\theta} = 0$, except where
\begin{eqnarray}
	\resampleu_t^i & = & c^{j}_t \mbox{ for $i, j=1,\ldots,N$},
\end{eqnarray}
then
\begin{eqnarray}
	\frac{d}{d\bm{\theta}} \mathbf{x}^{\prime i}_t & = & \frac{d}{d\bm{\theta}}x_t^{\kappa_i}
\end{eqnarray}
almost surely, such the derivative of the resampled state is obtained by taking the derivative of the parent particle.

\end{document}